\documentclass[conference]{IEEEtran}
\IEEEoverridecommandlockouts
\usepackage{cite}
\usepackage{amsmath,amssymb,amsfonts}
\usepackage{graphicx}
\usepackage{textcomp}
\usepackage{xcolor}
\usepackage{siunitx}
\usepackage{enumitem}
\usepackage[lined,ruled,linesnumbered]{algorithm2e}
\usepackage{gensymb}
\usepackage{caption}
\usepackage{subcaption}
\usepackage{multirow}

\def\BibTeX{{\rm B\kern-.05em{\sc i\kern-.025em b}\kern-.08em
    T\kern-.1667em\lower.7ex\hbox{E}\kern-.125emX}}
\begin{document}
\title{\textit{FEEL}: \underline{F}ast, \underline{E}nergy-\underline{E}fficient \underline{L}ocalization for Autonomous Indoor Vehicles}
%

\author{\IEEEauthorblockN{Vineet Gokhale\IEEEauthorrefmark{1},
Gerardo Moyers Barrera\IEEEauthorrefmark{2}, and
R. Venkatesha Prasad\IEEEauthorrefmark{3}}
\IEEEauthorblockA{Delft University of Technology, Delft, The Netherlands\\
Email: \IEEEauthorrefmark{1}V.Gokhale@tudelft.nl,
\IEEEauthorrefmark{2}G.I.MoyersBarrera@student.tudelft.nl,
\IEEEauthorrefmark{3}R.R.VenkateshaPrasad@tudelft.nl}}
\maketitle

\begin{abstract}
Autonomous vehicles have created a sensation in both outdoor and indoor applications. The famous indoor use-case is process automation inside a warehouse using Autonomous Indoor Vehicles (AIV). These vehicles need to locate themselves not only with an accuracy of a few centimeters but also within a few milliseconds in an energy-efficient manner. Due to these challenges, \textit{ localization} is a holy grail. In this paper, we propose \textit{FEEL} -- an indoor localization system that uses a fusion of three low-energy sensors: IMU, UWB, and radar. We provide detailed software and hardware architecture of FEEL. Further, we propose \textit{Adaptive Sensing Algorithm} (ASA) for opportunistically minimizing energy consumption of FEEL by adjusting the sensing frequency to the dynamics of the physical environment. Our extensive performance evaluation over diverse test settings reveal that FEEL provides a localization accuracy of $<$\SI{7}{cm} with ultra-low latency of $\approx$\SI{3}{ms}. Further, ASA yields up to 20\% energy saving with only a marginal trade off in accuracy.
\end{abstract}


\vspace{-5pt}
\section{Introduction}
\label{sec:intro}
In this paper, we propose \textit{FEEL} -- a fast, energy-efficient localization system using a sensor fusion of IMU, UWB, and Radar for Autonomous Indoor Vehicles (AIV). 

Recently, there have been many innovations and automation around the \textit{IoT, CPS, Industry 4.0, Autonomous vehicles \& drones}. These innovations are paving ways for a few hundreds or thousands of AIV to collaboratively execute complicated industrial processes efficiently. As an example, several AIVs could be executing all industrial tasks starting from supplying raw materials to a process in a factory to arranging finished goods in a warehouse. Efficiently performing these diverse tasks necessitate that the AIVs be fitted with sensors that can sense the operating environment and help the AIVs take intelligent decisions. A crucial part of this is \textit{localization} -- which is a \textit{holy grail} --  by which the AIVs learn their position in the physical environment. In a dense environment with several obstacles and narrow paths, localization accuracy of \SI{10}{cm} or lower is desired for safe operations. Furthermore, the localization should be achieved in a fast and energy-efficient manner.

\noindent\textbf{Why fast and energy-efficient?}\, Several applications of (semi-)autonomous robots requiring ultra-low latency ($<$\SI{10}{ms}) are fast-emerging. Primary examples include Tactile Internet \cite{holland2019} and autonomous drone racing \cite{cocoma2019}. Violation of latency requirements pose catastrophic consequences, such as crashes. This latency budget subsumes several operations like sensing, localization, path planning, communication, and actuation. In order to facilitate such applications with fast control loops, permissible localization latency is within a couple of milliseconds. In general, accurate and fast indoor localization can be achieved by employing powerful sensing and computational hardware (like LiDAR). However, this is not always an affordable solution since the AIVs are powered by batteries with limited energy budget in order to maintain a small form factor. Sub-optimal usage of energy results in the need for frequent recharging and replacement of batteries, thereby contributing heavily towards the overall carbon footprint.
Hence, achieving fast and energy-efficient localization without trading off accuracy is a challenging research problem.
\begin{figure}[!tbp]
    \centering
    \includegraphics[width=6.5cm, height=3.1cm]{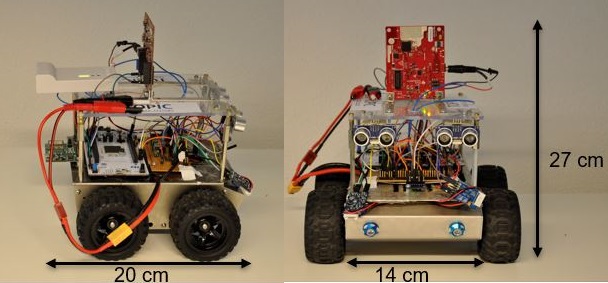}
\caption{Side and frontal views of FEEL -- the proposed fast, energy-efficient, and accurate localization system for AIV .}
\label{fig:robot}
\vspace{-0.6cm}
\end{figure}

For outdoor environments, Global Navigation Satellite System (GNSS) is the \textit{de facto} localization method \cite{nizhu2018}. However, due to its poor localization accuracy in indoor environments, several other sensors have been extensively investigated. While the vision-based sensors, like LiDAR and camera, can guarantee the required high accuracy for indoor environments, they are both energy-demanding (sensing and computation) as well as unreliable under poorly illuminated conditions. Hence, other sensors, such as radar, IMU, odometry, UWB, and WiFi have been extensively investigated in literature. Sensor fusion techniques are used to reap the benefits of simultaneously learning from multiple sensors.

Literature provides several combinations of these low-energy sensors fused in a variety of ways to maximize the localization accuracy. However, to the best of our knowledge, none of these combinations yields the necessary localization accuracy of $<$\SI{10}{cm}. Further, very less emphasis is put on the  characterization of latency and energy efficiency of these localization methods.
In this work, we attempt to bridge this gap. Our contributions in this work are as follows.

\noindent 1. We present the design of \textit{FEEL} -- a system that fuses three low-energy sensors -- IMU, UWB, and radar, for achieving fast and energy-efficient localization in the scale of a few centimeters in indoor environments. We design and develop a custom AIV for demonstrating the proof of concept of FEEL (see Figure~\ref{fig:robot}).

\noindent 2. We propose \textit{Adaptive Sensing Algorithm} (ASA) for opportunistically minimizing the energy consumption of FEEL by adjusting the sensing frequency to suit the dynamics of the operation environment. ASA offers the flexibility to tune its parameters to meet the application-specific energy-accuracy requirements.

\noindent 2. Our robust and extensive performance evaluation under diverse experimental settings reveal that FEEL can provide an accuracy of up to \SI{6.94}{cm} with an ultra-low latency of \SI{3.15}{ms}. Further, we show that ASA yields up to 20\% reduction in energy consumption by marginally trading off for accuracy. 
Comparing with state of the art in localization, we demonstrate the superiority and robustness of the performance of FEEL. 

\vspace{-3pt}
\section{Related Work} \vspace{-5pt}
\label{sec:related}
In this section, we will discuss only the recent works for providing a general view of the state of the art in localization.
Vision-based methods rely on at least one vision sensor, such as LiDAR and camera, for localization.
Zhen et al. employed LiDAR and UWB for localization in a tunnel-like environment  based on Error State Kalman Filter (ESKF) \cite{zhen2019}. Song et al. employed  LiDAR and RGB depth camera for localization using visual tracking and depth information \cite{song2016}. 
Wan et al. designed a robust localization method  through fusion of GNSS, LiDAR, and IMU \cite{wan2018}. 

Several localization methods using only non-vision based sensors, such as IMU, UWB, magnetometer, GNSS, and odometry, have also been extensively investigated. Hellmers et al. explored the combination of IMU and magnetometer
\cite{hellmers2018}. While the works in \cite{yao2017,marquez2017,feng2020} performed fusion between IMU and UWB, Dobrev et al. fused radar, ultrasound (US), and odometry data \cite{dobrev2016}. Haong et al. explored the potential of GNSS, infrared, and UWB \cite{hoang2017}, Zhou et al. developed a localization method using WiFi infrastructure \cite{zhou2019}. 

In general, vision-based sensors have shown the potential to yield a localization accuracy of sub-\SI{10}{cm}, however, they are both expensive (for ex., LiDAR costs a few thousand USD) and energy-intensive.
On the other hand, non-vision based sensors are generally cheap and consume low energy, albeit their localization accuracy is considerably low in comparison with vision-based systems. 
We present a system- and performance-level comparison of a few of the aforementioned works in Section~\ref{subsec:results}.
\vspace{-5pt}
\section{\textit{FEEL}: Design and Implementation}
\label{sec:proposed}
%
\subsection{Hardware Testbed}
\label{subsec:testbed}
The proper choice of sensors plays a vital role in determining the accuracy, latency, energy-efficiency, and cost of localization system. In this work, we explore the combination of inexpensive and low-energy sensors comprising of IMU, UWB, and radar. To the best of our knowledge, ours is the first attempt at localization using fusion of these three sensors. 

\noindent \textbf{IMU:} This is a cheap and low-power sensor that accurately captures acceleration and orientation of AIV. For our testbed, we employ MPU6050 \cite{IMU}. It offers a maximum sensing frequency of \SI{1}{kHz} and has an average power consumption of \SI{12.89}{mW}.

\noindent \textbf{UWB:} We use four UWB anchors placed at pre-defined location in the test environment and a tag placed on the AIV to obtain 2D position and velocity of the tag. For our testbed, we use Decawave DWM1001dev \cite{uwb} at a maximum sensing frequency of \SI{10}{Hz} and has an average power consumption of \SI{0.67}{W}. 

\noindent \textbf{Radar:} We employ radar for its robustness, ability to compensate for the erroneous measurements from UWB, and to detect objects in the vicinity of  AIV. In our testbed, the radar module used is AWR1843 which offers a maximum sensing frequency of \SI{130}{Hz} and has a average power consumption of \SI{1.92}{W}. 

The data processing and sensor fusion is performed on NUCLEO-L4R5ZI processor board \cite{processor}. It has an ARM Cortex M4 32-bit microcontroller
with an operating frequency of 120 MHz. The \SI{5}{V} DC motors for driving the wheels are powered using a Lipo 2S battery with a capacity \SI{3000}{mAh}.
\begin{figure}[!tbp]
    \centering
    \includegraphics[width=0.8\linewidth]{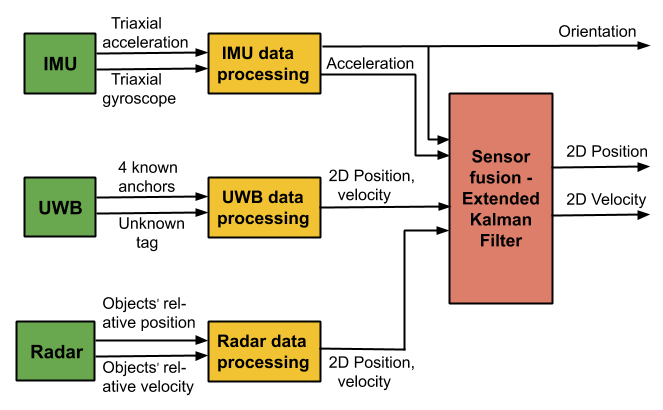}
\caption{Block diagram representation of FEEL showing the sensor fusion of IMU, UWB, and radar.}
\label{fig:testbed}
\vspace{-0.6cm}
\end{figure}
\vspace{-5pt}
\subsection{Sensor Fusion}
\label{subsec:fusion}
In this work, we use Extended Kalman Filter (EKF) as the sensor fusion technique. For simplicity in implementation, we use IMU data as the control input to the predict position and velocity of EKF, whereas the IMU orientation completely determines the AIV orientation as shown in Figure~\ref{fig:testbed}. We now provide a detailed description of the different steps of EKF in the design of FEEL.

\noindent\textbf{Prediction:} Let $(a_x, a_y)$ denote the 2D acceleration measured by IMU and $\theta$ denote IMU orientation with respect to the initial position. Let $(x, y) \text{ and } (v_x,v_y)$ be the state variables of EKF denoting 2D position and velocity of AIV, respectively. Denoting the state variables as $X=[x \phantom{a} y\phantom{a} v_x\phantom{a} v_y]^T$, their prediction  $\hat{X}$ at time step $k$ is then expressed as, 
\begin{equation*}
\hat{X_k} = FX_{k-1} + Bu_k + w_k,
\label{eq:prediction}
\end{equation*}
where\\
$$
F = 
 \begin{bmatrix}
1 & 0 & \Delta t & 0 \\
0 & 1 & 0 & \Delta t  \\
0 & 0 & 1 & 0 \\
0 & 0 & 0 & 1
\end{bmatrix},
B = \begin{bmatrix}
\frac{1}{2} \Delta t^{2} &0 &0 &0\\
0 & \frac{1}{2} \Delta t^{2} & 0 & 0\\ 
0 & 0&  \Delta t & 0\\
0 & 0&  0 & \Delta t \\
\end{bmatrix},
$$\\
$u_k = [a^k_x\phantom{a} a^k_y\phantom{a} a^k\text{sin}\theta^k\phantom{a} a^k\text{cos}\theta^k]^T$ is the control input, and $a$ is the magnitude of acceleration.
Here, $F \text{ and } B$ denote the state transition matrix and control matrix, respectively, $\Delta t$ is the time between successive measurements of IMU, and $w_k\sim N(0, Q)$ denotes the measurement noise of IMU. Note that we use subscripts for matrices and superscripts for state variables to indicate the time step $k$. The noise covariance of prediction is given by 
\begin{equation*}
\hat{P_k} = FP_{k-1}F^T + Q,
\end{equation*}
where $Q$ is the noise covariance of $u$ expressed as
$$
 Q   =  \begin{bmatrix}
\sigma_{a_x}^{2} & 0 & 0 & 0 \\
0 & \sigma_{a_y}^{2} & 0 & 0  \\
0 & 0 & \sigma_{sin\theta}^{2} & 0 \\
0 & 0 & 0 & \sigma_{cos\theta}^{2}
\end{bmatrix}
$$

\vspace{1mm}
\noindent\textbf{Measurement:} This step relies on the position and velocity measurements from UWB and radar. Due to the dependence on two different sensors measurement, we will employ a weighted averaging to determine the effective measurement from them. The measurement matrix $Z_k$ is given as
$$
Z_k =\begin{bmatrix}
{\bar{x}^k} \\
{\bar{y}^k}\\ 
{\bar{{v}}_x}^k \\
{\bar{{v}}_y}^k \\
\end{bmatrix} + n_k = \begin{bmatrix}
x^k_u\alpha_x + x^k_r(1-\alpha_x) \\
y^k_u\alpha_y + y^k_r(1-\alpha_y)\\ 
v^k_{x(u)}\beta_x + v^k_{x(r)}(1-\beta_x)\\
v^k_{y(u)}\beta_y + v^k_{y(r)}(1-\beta_y) \\
\end{bmatrix}+n_k,
$$
where $(x_u,y_u) \text{ and } (v_{x(u)},v_{y(u)})$ are the 2D position and velocity measurements of UWB, respectively. Similarly,  $(x_r,y_r) \text{ and } (v_{x(r)},v_{y(r)})$ are the 2D position and velocity measurements of radar, respectively. $\alpha_x \text{ and } \alpha_y$ denote weights assigned to the position measurements of UWB along $x$ and $y$-axis, respectively. Similarly, $\beta_x \text{ and } \beta_y$ denote weights assigned to the velocity measurements of UWB along $x$ and $y$-axis, respectively. The weighted average of the sensor measurements is denoted with the overline character. Since the weighted average is used as state variables in the measurement model, $H$, the Jacobian of $Z$, is an identity matrix, i.e. $H=I$. The measurement noise $n_k \sim N(0, R)$, where
$$
R = \begin{bmatrix}
\sigma^2_{\bar{x}} & 0 & 0 & 0 \\
0 & \sigma^2_{\bar{y}} & 0 & 0  \\
0 & 0 &  \sigma^2_{\bar{{v}}_x} & 0 \\
0 & 0 & 0 & \sigma^2_{\bar{{v}}_y}
\end{bmatrix}.
$$
Here $\sigma^2_{\bar{x}} = \alpha_x\sigma_{x_u}^{2} + (1-\alpha_x)\sigma_{x_r}^{2}$, $\sigma^2_{\bar{y}} = \alpha_y\sigma_{y_u}^{2} + (1-\alpha_y)\sigma_{y_r}^{2}$, $\sigma^2_{\bar{{v}}_x} = \beta_x\sigma_{v_{x(u)}}^{2} + (1-\beta_x)\sigma_{v_{x(r)}}^{2}$, and   $\sigma^2_{\bar{{v}}_y} = \beta_y\sigma_{v_{y(u)}}^{2} + (1-\beta_y)\sigma_{v_{y(r)}}^{2}$. 

\vspace{1mm}
\noindent\textbf{Update:} After the measurement values are available, the EKF applies the correction to the predicted values as per the following equations. The innovation matrix $Y$, its covariance $S$, and Kalman gain $K$ are given by the following equations.
\begin{equation*}
    Y_k = Z_k-\hat{X_k}, \phantom{a}S_k=\hat{P_k}+R, \phantom{a} K_k=\hat{P_k}S_k^{-1}
\end{equation*}
Finally, the prediction and its covariance matrices are updated as
\begin{equation*}
    X_k = \hat{X_k}+K_kY_k, \phantom{a} P_k=(I-K_k)\hat{P_k}.
\end{equation*}
Due to the need of accurate localization as well as reducing computational demands of FEEL, we run EKF in synchronization with IMU at \SI{1}{kHz}. This means that UWB-radar samples received in between IMU samples are processed only after next IMU sample arrives.

Having presented the hardware and software design, we now move to enhancing FEEL by making it energy-efficient through the design of ASA. 
\vspace{-3pt}
\subsection{Adaptive Sensing Algorithm (ASA)}
\label{subsec:adaptive}
As mentioned in Section~\ref{subsec:testbed}, the power consumption of sensors used in FEEL is substantially lower than LiDAR-based localization systems (whose exact values are presented later in Table~\ref{table:related}). However, as a good engineering practice, it is important that the AIV utilizes only as much power as required to meet the different demands of any given application. This improves the energy demands of AIV and eventually contributes towards development of eco-friendly industries. This motivates us to design \textit{Adaptive Sensing Algorithm} (ASA) to improve the energy-efficiency of FEEL. The idea behind ASA is to opportunistically sense information depending on the nature of AIV trajectory and environment. This is done by adjusting the sampling frequency of sensors according to the changes in the environment such that the sensing energy consumption is minimized without trading off much on the localization accuracy.
\begin{algorithm}[!h]
\caption{Adaptive Sensing Algorithm}
\label{alg:asa}
\tcc{For radar and UWB, add/prefix subscript $r$ and $u$, respectively, to $f,f_{(min)},f_{(max)},\gamma,m,c$}
\SetAlgoLined
\vspace{2mm}
\KwResult{Determine $f$ based on $\delta\theta$ and $d$}
 Set $m_u=1, m_r=0.5,c_u=-1,c_r=0$
 
 Initialize: $f= f_{max}$

    \uIf{$\delta\theta<\theta'$}{
            \uIf{$f_{(min)}<f\leq\gamma$}
            {
                $f\gets mf+c$
            }
        \uElseIf{$f=f_{(max)}$}{
                $f\gets\gamma$
            }
        }
    \Else{
        \uIf{$f_{(min)}\leq f<\gamma$}
        {
            $f \gets\gamma$    
        }
        \Else{$f\gets f_{max}$}
    
    }

  \uIf{$\phantom{a} d<d'$}
  {
        $f_r\gets f_{r(max)}$
  }
  Wait for time $T$ and go back to Step 3
\end{algorithm}

As already discussed in Section~\ref{subsec:testbed}, the energy consumption of IMU is significantly lower than UWB and radar. Hence, in ASA we will maintain IMU sampling frequency at its maximum (\SI{1}{kHz}) while adapting only the sampling frequency of UWB and radar, denoted by $f_u \text{ and } f_r$, respectively. 
Naturally, the power consumption (localization error) monotonically increases (decreases) with sampling frequency, although characterizing the exact relationship is non-trivial. For simplicity, we define \textit{threshold frequencies} $\gamma_u \text{ and } \gamma_r$ as the minimum $f_u \text{ and } f_r$, respectively, such that there is negligible reduction in accuracy ($<$\SI{1}{cm}) in comparison with that at respective maximum sampling frequency. Let us denote the permissible minimum  and maximum frequency of UWB as $f_{u(min)} \text{ and } f_{u(max)}$, respectively, and those of radar as $f_{r(min)} \text{ and } f_{r(max)}$, respectively. Note that $\gamma_u \in (f_{u(min)}, f_{u(max)})$ and $\gamma_r \in (f_{r(min)}, f_{r(max)})$.
%
%
%

ASA works by increasing the sampling frequency in two scenarios:
\begin{enumerate}[leftmargin=*]
    \item Trajectory (orientation) of AIV changes significantly: If $\delta\theta>\theta'$, then the localization of both UWB and radar is impacted heavily. Here, $\delta\theta$ denotes the change in $\theta$ per time duration $T$ and $\theta'$ is the orientation threshold used by ASA. In this case, both $f_u \text{ and } f_r$ are increased aggressively to minimize the localization error. The frequency is increased as follows: If $f_r=\gamma_r$, then $f_r\gets f_{r_(max)}$, and if $f_{r(min)} \leq f_r < \gamma_r$, then $f_r\gets\gamma_r$. Adaptation of $f_u$ also happens in a similar fashion.
    \item External objects in close proximity of AIV: If $d\leq d'$,  then only localization of radar is impacted. Here $d$ and $d'$ denote the distance of AIV measured by radar and threshold used by ASA, respectively. In this case, $f_r\gets f_{r(max)}$ in order to minimize localization error of radar.
\end{enumerate}
If none of the above scenarios occur, then localization accuracy can be traded off slightly for obtaining energy saving. To achieve this, ASA prudently reduces $f_u \text{ and } f_r$ until $f_{u(min)} \text{ and } f_{r(min)}$ are reached, respectively. Due to the heterogeneous ranges of sampling frequencies, i.e. $f_u\in [1,10]\text{ Hz} \text{ and } f_r\in [1,130]\text{ Hz}$, we choose to reduce $f_u$ linearly and $f_r$ multiplicatively.  
The working of ASA is presented in Algorithm~\ref{alg:asa}.

The efficacy of ASA lies in the fact that the user can configure its parameters to obtain the desired application-specific energy-accuracy performance from FEEL.

\section{Performance Evaluation}
\label{sec:performance}
\begin{figure*}[!tbp]
\centering
  \begin{tabular}[b]{c}
    \includegraphics[width=0.3\textwidth]{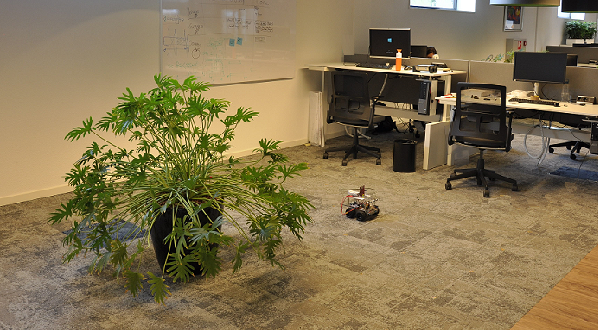} \\
    \small (a)
  \end{tabular} 
  \begin{tabular}[b]{c}
    \includegraphics[width=0.3\textwidth]{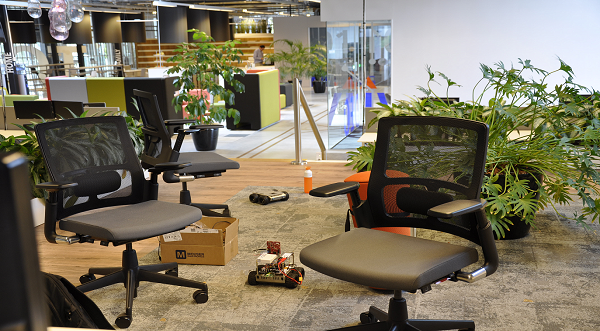} \\  
    \small (b)
  \end{tabular}
  \begin{tabular}[b]{c}
   \includegraphics[width=0.3\textwidth]{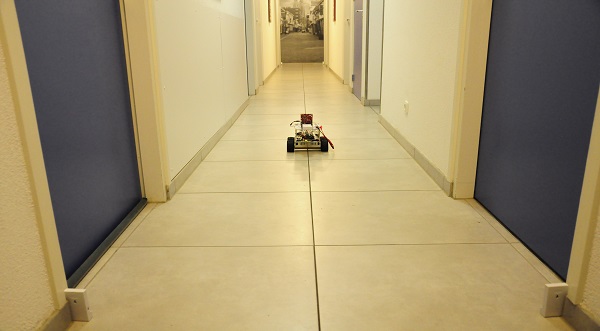} \\  
    \small (c)
  \end{tabular}
\caption{Indoor test environments used for performance evaluation of FEEL. a) E$_1$: office environment with sparsely distributed objects, b) E$_2$: office environment with sparsely distributed objects, and c) E$_3$: narrow, long corridor with no other objects.}
    \label{fig:environment}
\end{figure*}
We now move to comprehensive performance evaluation of FEEL where we will first describe the experimental setup  (Section~\ref{subsec:setup}) and then shed light on the important findings of our evaluation (Section~\ref{subsec:results}). 
\subsection{Experimental Setup}
\label{subsec:setup}
In order to conduct a robust performance evaluation under a wide variety of operating conditions, we consider different profiles of indoor test environment, movement track, and vehicle speed.

\vspace{1mm}
\noindent\textbf{Environment profile:} We evaluate FEEL in three test environments with different physical characteristics as shown in Figure~\ref{fig:environment}.
\begin{description}[leftmargin=*]
   \item \textbf{E$_1$:} In this environment, a couple of large objects are located around the track as shown in Figure~\ref{fig:environment}a. E$_1$ enables us to understand the performance of FEEL in a typical office environment with sparse distribution of objects. 
   \item \textbf{E$_2$:} In this environment,  there are several objects, such as chairs, tables, and a few plants, located in close proximity to the track as shown in Figure~\ref{fig:environment}b. E$_2$ enables us to understand the performance of FEEL in a typical factory/warehouse environment with dense distribution of objects. In our experiments, the locations of E$_2$ and E$_1$ are the same. 
   \item \textbf{E$_3$:} This is a narrow, long corridor with no objects in it apart from AIV as shown in Figure~\ref{fig:environment}c. E$_3$ enables us to understand the performance of FEEL in narrow spaces such as tunnels with extremely low margin of error and strong radio wave reflections off the walls.
\end{description}

\vspace{1mm}
\noindent\textbf{Track profiles:} In our investigation, we consider two track profiles: \textit{straight track} and \textit{race track}. As the names suggests, while in straight track profile the AIV drives along a straight line between two pre-defined points, in race track profile, the AIV trajectory resembles an oval shape. 

\vspace{1mm}
\noindent\textbf{Speed profiles:} Experimenting with different speed profiles helps us identify the limits of AIV speeds for which FEEL can provide necessary performance guarantees. Since the maximum possible speed of our AIV is \SI{4}{kmph}, we investigate the performance of FEEL with two speed profiles: low (\SI{1.2}{kmph}) and high (\SI{4}{kmph}). 

We conduct extensive experimentation with all combinations of the above profiles except for race track in E$_3$, since E$_3$ has no scope for bends and turns because of its narrow width. 

As mentioned in Section~\ref{subsec:fusion}, EKF runs at \SI{1}{kHz} implying that $\Delta t$=\SI{1}{ms}. In order to compute $Q \text{ and } R$, we record sensor measurements by running several tests under different experimental conditions described above. By comparing the recorded values against ground truth, we deduce $[\sigma^2_{a_x}, \sigma^2_{a_y}, \sigma^2_{\text{sin}\theta}, \sigma^2_{\text{cos}\theta}]$=[2.31, 0.60, 0.32, 0.65]$\times10^{-3}$. By empirically setting $\alpha_x$=$\alpha_y$=0.7 and $\beta_x$=$\beta_y$=0.4, we obtain $[\sigma^2_{\bar{x}}, \sigma^2_{\bar{y}}, \sigma^2_{\bar{{v}}_x}, \sigma^2_{\bar{{v}}_y}]$=[0.14,0.06, 0.13,0.11]. As concerns the parameters of ASA, we empirically set $\theta'$=\SI{10}{\degree}, $d'$=\SI{1}{m}, and $T$=\SI{1}{s}. These parameters can be tuned to suit the performance of FEEL to application-specific requirements. Additionally, $f_{u(max)}$=\SI{10}{Hz} and $f_{r(max)}$=\SI{130}{Hz}. Other parameter configurations, such as $f_{u(min)}, f_{r(min)}, \gamma_u, \text{ and } \gamma_r$, will be discussed in the next section.
\subsection{Experimental Results}
\label{subsec:results}
\begin{figure*}[!tbp]
\centering
  \begin{tabular}[b]{c}
    \includegraphics[width=5.4cm, height=3.4cm]{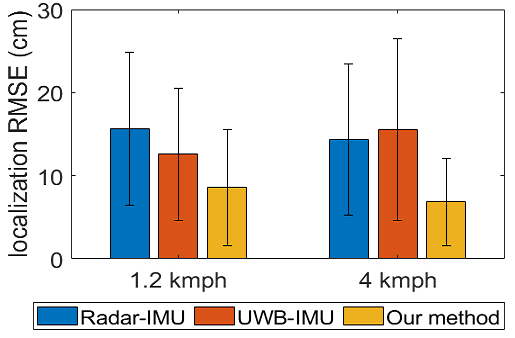} \\
    \footnotesize (a)
  \end{tabular} \hspace{-5mm}
  \begin{tabular}[b]{c}
    \includegraphics[width=5.4cm, height=3.4cm]{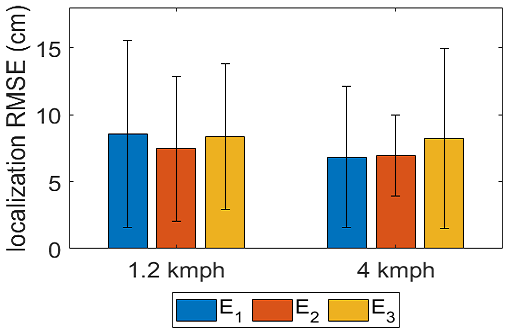} \\  
    \footnotesize (b) 
  \end{tabular} \hspace{-5mm}
  \begin{tabular}[b]{c}
   \includegraphics[width=5.6cm, height=3.4cm]{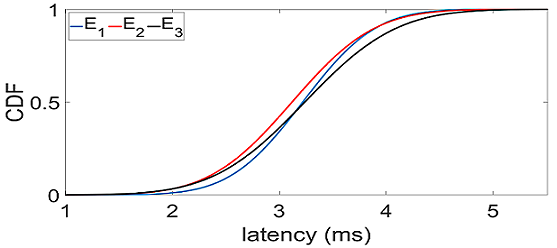} \\  
    \footnotesize (c) 
  \end{tabular}
\caption{Performance evaluation of FEEL. (a) Comparison of localization RMSE with two common localization methods, (b) Localization RMSE in three test environments (c) Localization latency in three environments and high speed profile.}
    \label{fig:error-latency}
    \vspace{-4mm}
\end{figure*}



We begin by presenting the localization accuracy and latency performance of FEEL. 
Note that although we conduct performance evaluation across several profiles explained earlier, due to space constraints we present only the most interesting observations in this paper.   

Figure~\ref{fig:error-latency}a shows a head-to-head comparison of localization accuracy of FEEL against two commonly used non-vision based localization techniques: IMU-radar and IMU-UWB. In order to obtain a fair and consistent performance analysis, we implement these localization techniques on our testbed. The presented findings correspond to the two speed profiles in E$_1$ and straight track. It can be seen that FEEL outperforms others giving RMSE of \SI{8.57}{cm} and \SI{6.85}{cm} for low and high speed profiles, respectively. FEEL yields a significant accuracy improvement of up to 2x and 2.2x over IMU-radar and IMU-UWB, respectively. Interestingly, the accuracy of our method increases at high speed. This behavior is attributed to the localization component of radar module used in our testbed, since its accuracy is known to increase with speed.

We present the localization accuracy of FEEL in the three test environments for straight track and both speed profiles in Figure~\ref{fig:error-latency}b. It can be seen that in all cases, the RMSE is sub-\SI{9}{cm}. Interestingly, for low speed profile, the performance in E$_2$ better than in E$_1$ and E$_3$. The rationale behind this behavior is that the large number of objects present in E$_2$ aid radar in localizing the AIV better. We observe comparable performance of FEEL in case of race track profile also.     

We now move to measuring the localization latency of FEEL. Figure~\ref{fig:error-latency}c presents our findings for straight track in the three test environments. It can be observed that the mean latency is around \SI{3.15}{ms}. The extremely low latency profile demonstrates the potential of the usage of FEEL in ultra-low latency applications. Further, FEEL is also robust in its latency performance to the characteristics of operating environment.     
\begin{table}[!h]
\large
\centering
\resizebox{0.5\textwidth}{!}{
\begin{tabular}{|c|c|c|c|c|c|}
\hline
 & IMU-UWB & UWB-LiDAR & IMU-UWB- & Radar-US & IMU-UWB \\
 & \cite{marquez2017} & \cite{song2019} & LiDAR \cite{zhen2019} & \cite{dobrev2016} & -Radar (Ours) \\ \hline
\Large Test speed & \Large\multirow{2}{*}{2.5} & \Large\multirow{2}{*}{2.88} & \Large\multirow{2}{*}{2.52} & \Large\multirow{2}{*}{4.32} & \Large\multirow{2}{*}{4} \\
(kmph) & & & & & \\ \hline  
\Large Max. accuracy & \Large\multirow{2}{*}{10.2} & \Large\multirow{2}{*}{7.6} & \Large\multirow{2}{*}{10} & \Large\multirow{2}{*}{15} & \Large\multirow{2}{*}{6.94} \\ 
(cm) & & & & & \\ \hline
\Large Mean latency & \Large\multirow{2}{*}{--} & \Large\multirow{2}{*}{3} & \Large\multirow{2}{*}{--} & \Large\multirow{2}{*}{--} & \Large\multirow{2}{*}{3.15} \\ 
(ms) & & & & & \\ \hline
\Large Power consp. & \Large\multirow{2}{*}{--} & \Large\multirow{2}{*}{$>$10} & \Large\multirow{2}{*}{$>$10} & \Large\multirow{2}{*}{$\sim$6.06} & \Large\multirow{2}{*}{4.19} \\ 
(W) & & & & & \\ \hline
 \Large System cost & \Large\multirow{2}{*}{--} & \Large\multirow{2}{*}{$>$4500} & \Large\multirow{2}{*}{$>$4500} & \Large\multirow{2}{*}{$>$4000} & \Large\multirow{2}{*}{$\sim$400} \\ 
(\$) & & & & & \\ \hline
\end{tabular}
}
\caption{System- and performance-level comparison of FEEL with state of the art in localization.}
\label{table:related}
\end{table}

We now compare the performance of FEEL (without ASA) with state of the art in localization. Note that the power consumption presented is for the overall system including the sensors. As can be seen from Table~\ref{table:related}, only UWB-LiDAR \cite{song2019} matches the accuracy and latency provided by FEEL, however this comes at the expense of high power consumption and overall system cost since it employs LiDAR. 
The missing details of \cite{marquez2017} is due to the insufficient details provided in the article. 
To summarize, FEEL outperforms the state of the art in localization comprehensively.   

\begin{figure}[!t]
    \centering
    \includegraphics[width=7.6cm, height=3.9cm]{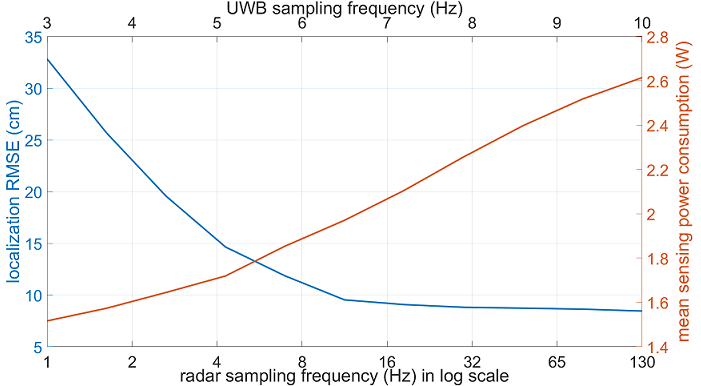}
\caption{Trade off between localization error and sensing power consumption over a wide range of $f_r \text{ and } f_u$ in E$_2$.}
\label{fig:errorpower}
\vspace{-0.5cm}
\end{figure}

We now move to the energy-efficiency part of FEEL. Before assessing the performance of ASA, it is important to understand the influence of $f_r \text{ and } f_u$ on localization accuracy and power consumption in order to determine $f_{r(min)}, f_{u(min)}, \gamma_r \text{ and } \gamma_u$. Figure~\ref{fig:errorpower} presents these dynamics for E$_2$ and race track profile. It can be seen that while the localization error falls off exponentially with increasing $f_r \text{ and } f_u$, the sensing power consumption increases linearly on a log scale for $f_r$ and a linear scale for $f_u$. Interestingly, radar information has significant redundancy that can be exploited to achieve energy efficiency in ASA. It is worth mentioning that the variation looks similar for other experimental profiles also. From Figure~\ref{fig:errorpower}, it is clear that $\gamma_r$=\SI{16}{Hz} and $\gamma_r$=\SI{7}{Hz}. It can be seen that at these values, there is $\approx18\%$ power saving in comparison with that at maximum frequency. This demonstrates the scope of improving the energy efficiency of FEEL with negligible compromise on accuracy. For our experiments, we fix $f_{r(min)}$= \SI{4}{Hz} and $f_{u(min)}$=\SI{5}{Hz}.

\begin{figure}[!t]
    \centering
    \includegraphics[width=8.5cm, height=4.7cm]{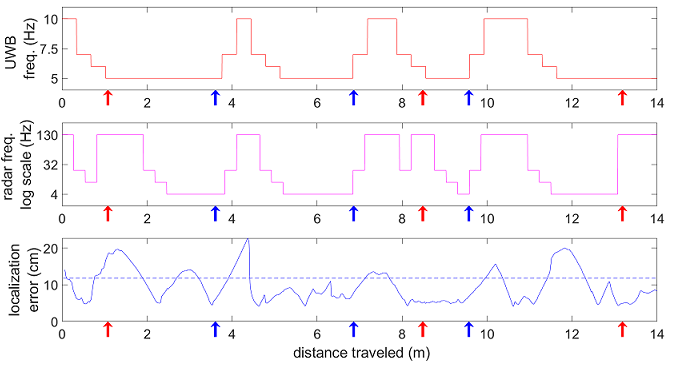}
\caption{Demonstration of sampling frequency adaptation of ASA and the resulting localization error in response to turns (blue arrows) and objects (red arrows) in the environment. }
\label{fig:adaptive}
\vspace{-0.4cm}
\end{figure}

With the above values for frequency parameters, we now present the performance evaluation of ASA in E$_1$ on race track. The total length of the track is \SI{14}{m} with three large objects around the track. The AIV starts at a corner and stops at the same without making the last turn. The three objects and turns are marked in Figure~\ref{fig:adaptive} using red and blue arrows, respectively. As can be seen, frequency increase as a response to objects starts prior to reaching the object due to the radar detecting them, whereas response to turns starts only once the AIV starts changing its trajectory.
Due to this adaptation by ASA, there is a $\approx20\%$ reduction in power of AIV, although the localization RMSE is measured to be \SI{12}{cm}. Note that although we worked with empirical choices of $f_{r(min)} \text{ and } f_{u(min)}$, ASA provides the flexibility to configure these values to obtain the application-specific energy-accuracy performance.  
\section{Conclusions}
Autonomous indoor vehicles will be heavily employed in automation industries/warehouses in the near future. These AIVs require high localization accuracy at ultra-low latency and low-energy demands. In this paper, we proposed a system called \textit{FEEL} for indoor localization by fusing IMU, UWB, and radar. We presented the design of the custom-built hardware for showing the proof of concept. In order to address the energy challenges, we proposed \textit{Adaptive Sensing Algorithm (ASA)} for opportunistically tuning the sampling frequency to minimize energy consumption by trading off marginally for accuracy. We performed extensive evaluation and found that the position accuracy of FEEL is within \SI{7}{cm} while the latency is around \SI{3}{ms}. We showed that ASA provides a considerable improvement in the energy performance of FEEL of up to 20\%. Further, we showed that FEEL comprehensively outperforms state of the art in indoor localization. We believe that the work presented here will become one of the benchmarks for further research in this domain.   
\label{sec:conclusions}

\bibliographystyle{IEEEtran}
\bibliography{refs}

\end{document}